\documentclass[10pt,twocolumn,letterpaper]{article}

\usepackage{ijcb}
\usepackage{times}
\usepackage{epsfig}
\usepackage{graphicx}
\usepackage{amsmath}
\usepackage{amssymb}
\usepackage{comment}
\usepackage{import}
\usepackage{enumitem}
\usepackage{soul}
\usepackage{hyperref}
\usepackage{balance}
\usepackage{bm}
\usepackage{xcolor,colortbl}
\usepackage{subcaption}

\definecolor{boxgrey}{HTML}{F3F3F3}
\newcolumntype{a}{>{\columncolor{boxgrey}}r}

\newcommand{\hlbox}[2]{
  \begin{center}
    \fcolorbox{white}{boxgrey}{
      \parbox{0.9\columnwidth}{\noindent \textbf{#1}. \textit{#2}}
    }
  \end{center}
}
\ijcbfinalcopy

\ifijcbfinal\fi
\begin{document}

\title{Explaining Bias in Deep Face Recognition via Image Characteristics}

\author{Andrea Atzori, Gianni Fenu, Mirko Marras\\
Department of Mathematics and Computer Science, University of Cagliari, Cagliari, Italy\\
{\tt\small andrea.atzori@unica.it, fenu@unica.it, mirko.marras@acm.org}}

\maketitle

\thispagestyle{empty}

\begin{abstract}
In this paper, we propose a novel explanatory framework aimed to provide a better understanding of how face recognition models perform as the underlying data characteristics (protected attributes: gender, ethnicity, age; non-protected attributes: facial hair, makeup, accessories, face orientation and occlusion, image distortion, emotions) on which they are tested change. With our framework, we evaluate ten state-of-the-art face recognition models, comparing their fairness in terms of security and usability on two data sets, involving six groups based on gender and ethnicity. We then analyze the impact of image characteristics on models performance. Our results show that trends appearing in a single-attribute analysis disappear or reverse when multi-attribute groups are considered, and that performance disparities are also related to non-protected attributes. Source code: \url{https://cutt.ly/2XwRLiA}. 
\end{abstract}

\section{Introduction}

\vspace{1mm} \textbf{Motivation}. 
Face biometrics is deployed in several applications, such as access control, social media information structuring, or media archive organization.
Performance has strongly progressed in recent years due to the introduction of deep learning techniques~\cite{wang2021deep}. 
Results reported for public data sets from heterogeneous sources are getting close to 100\% accuracy~\cite{wang2021facex}.
However, biometric systems based on face recognition are exhibiting demographic biases that lead to unfair treatments~\cite{mitigate_racial,genderineq,sexist}. 
For example, face detection methods failed more often on subjects with darker skin-tone than those with lighter skin-tone~\cite{8638319,9186002}. 
Face recognition models performed better on Caucasian faces than East Asian faces~\cite{phillips2011other}. 

\vspace{1mm} \textbf{Problem Statement}. 
The increasing attention to these biases brings into question the fairness and integrity of systems empowered with face biometrics. Fairness is a concept of \emph{non-discrimination} on the basis of the membership to \emph{protected groups}, identified by a \emph{protected attribute}, e.g., gender and age in anti-discrimination legislation\footnote{Please consult Art. 21 EU Charter of Fundamental Rights, Art. 14 EU Convention on Human Rights, Art. 18-25 EU Treaty on the Functioning.}. \emph{Group fairness} avoids the discrimination of a given group, assessed as the absence of a \emph{disparate impact} in the outcomes generated for its members~\cite{DBLP:journals/csur/MehrabiMSLG21}. Group fairness should therefore account for no disparate impact of error rates on protected groups of individuals. Providing guarantees on this property is essential for the responsible adoption of face biometrics.

\vspace{1mm} \textbf{Open Issues}. 
Prior work has showed that the error rates are usually higher for certain demographic groups and mitigated this bias by attempting to make error rates between individuals belonging to different protected groups statistically indistinguishable, e.g.,~\cite{FairFace,FairnessFaceMatch,mitigate_racial}. However, these analyses were run on one sensitive attribute at time. Focusing on groups characterized by a sensitive attribute does not necessarily provide guarantees and generalization to groups based on another sensitive attribute~\cite{kleinberg2016inherent}. Notable effects of this operation mode are observed when a trend appears in different groups of observational data but disappears or even reverses when these groups are combined. Our first key aim in this paper is to provide a robust understanding to this issue, under a multiple protected attribute perspective.  

\begin{table*}
\resizebox{\textwidth}{!}{%
\begin{tabular}{ll|lllll|llllll|lllll}
\hline
\multicolumn{2}{l}{\textbf{Study Information}}               & \multicolumn{5}{|l}{\textbf{Sensitive Attribute(s)}$^1$} & \multicolumn{6}{|l|}{\textbf{Face Data Set(s)}} & \multicolumn{5}{|l}{\textbf{Evaluation Metric(s)$^2$}}          \\
\textit{Authors}                                             & \textit{Type}       & \textit{G}    & \textit{E}    & \textit{A}    & \textit{S}   & \textit{I}   & \textit{VGG} & \textit{IJB} & \textit{LFW} & \textit{BUPT} & \textit{RFW} & \textit{Other(s)}                  & \textit{FPR/FNR} & \textit{ROC} & \textit{AUC} & \textit{Accuracy} & \textit{Other(s)} \\
\hline
Terhorst, P et al. \cite{comp_demo}                 & Analysis   & X    & X    &      & X   &     &     &           &     &      &     & MAAD                         & X       &     &     &          & EER, MAD      \\
Majumdar et al \cite{unravelling}                   & Analysis   & X    & X    &      &     &     &     &           &     &      &     & MORPH, MUCT                  &         &     &     &          & DoB      \\
Albiero et al. \cite{genderineq}                    & Analysis   & X    & X    &      &     &     &     &           &     &      &     & MORPH                        & X       & X   &     &          &          \\
Albiero \& Bowyer \cite{sexist}                      & Analysis   & X    &      &      &     &     &     &           &     &      &     & MORPH, Notre Dame            & X       &     &     &          &          \\
Dooley et al. \cite{robust_face_det}                & Analysis   & X    & X    & X    &     &     &     &           &     &      &     & O-Images, MIAP, CCD, UtkFace &         &     &     &          & mrCE     \\
Hupont \& Fernandez \cite{demogpairs}               & Analysis   &      & X    &      &     &     & X   & X         & X   &      &     & CWF                          & X       &     &     &          &          \\
Popescu et al. \cite{popescu2021face}               & Analysis   & X    & X    & X    &     &     &     & X         & X   &      &     &                              &         &     &     & X        &          \\
Rosenberg et al. \cite{FaceRecObf}                  & Analysis   & X    & X    &      &     &     & X   &           & X   &      &     &                              & X       &     & X   &          &          \\
\hline
Kortylewskiet et al. \cite{kortylewski2019synthetic}& Mitigation &      &      &      &     & X   &     & X         & X   &      &     & CMU                          & X       &     &     &          &          \\
Gwilliam et al. \cite{mitigate_racial}              & Mitigation & X    & X    &      &     &     &     &           &     & X    & X   &                              &         &     &     & X        &          \\
Wang et al. \cite{9512390}                          & Mitigation &      & X    & X    & X   &     &     &           &     & X    &     &                              &         & X   &     & X        &          \\
Alasadi et al. \cite{FairnessFaceMatch}             & Mitigation &      & X    &      &     &     &     &           &     &      &     & Celeb-A/UMD                  & X       &     &     & X        &          \\
Wang et al. \cite{Wang2019RacialFI}                 & Mitigation &      & X    &      & X   &     &     & X         &     &      & X   &                              & X       &     &     & X        &          \\
Gong et al. \cite{groupadaptivemitig}               & Mitigation &      & X    &      &     &     &     & X         & X   & X    & X   &                              & X       &     &     & X        &          \\
Rodriguez et al. \cite{9025435}                     & Mitigation & X    &      &      &     &     & X   &           &     &      &     &                              & X       & X   & X   &          &          \\
Terhorst, P et al. \cite{comparisonlevelmitig}      & Mitigation &      & X    &      &     &     &     &           & X   &      &     &                              & X       & X   & X   &          & EER      \\
\hline
\multicolumn{18}{l}{$^1$ G : Gender, E : Ethnicity, A: Age, S: Soft Attribute(s), I : Identity.} \tabularnewline
\multicolumn{18}{l}{$^2$ FPR/FNR : False Positive/Negative Rate, AUC : Area under the ROC Curve, EER : Equal Error Rate, DoB : Degree of Bias, mrCE : mean relative Corruption Error, MAD : Mean Absolute Deviation.} \tabularnewline
\end{tabular}

}
\vspace{-4mm}
\caption{Summary table about research on exploration and mitigation of unfairness in face recognition.}
\label{tab:prior-work}
\end{table*}

This issue directly brings forth a discussion on better understanding \emph{why} a face recognition model may lead to a different performance for different (groups of) individuals. It can be even questioned that the reasons behind a disparate impact are directly related to a (protected) attribute, and not to other covariates. Existing studies on face recognition rarely investigated potential sources of unfairness, beyond the under-representation of certain demographic groups in the data~\cite{inbook,9512390,9025435,demogpairs}. Inspired by recent studies going beyond demographics~\cite{terhorst2021comprehensive}, our second key aim in this paper is to dig deeper into the potential influence of certain data characteristics on the security and usability levels achieved by the face recognition model and, consequently, on the disparate impacts among demographic groups.

\vspace{1mm} \textbf{Our Contributions}. 
In view of the aforementioned objectives, in this paper, we propose a framework for understanding how face recognition models perform as the underlying data characteristics (protected and non-protected attributes) on which they are tested change, considering security, usability, and fairness. Our framework is based on a statistical model that involves over 20 data characteristics and studies their impact on the performance of ten state-of-the-art face recognition models. This impact is referred to as \emph{explanatory power}, as it reflects the capability of the resulting characteristic to influence a dependent variable (either security or usability). Specifically, we aim to address the following research questions: do unfairness phenomena observed in face recognition under a single protected attribute change when we consider multiple protected attributes simultaneously (\emph{RQ1}), is there any co-relationship between image characteristics and error rates (\emph{RQ2}), how a change in an image characteristic impacts error rates (\emph{RQ3}). To answer these questions:

\begin{itemize}[leftmargin=5mm,itemsep=1pt]
\item We present a novel regression-based explanatory framework developed to analyze the impact of data characteristics on fairness estimates experienced by face recognition models, targeting security and usability dimensions.
\item We evaluate ten state-of-the-art face recognition models, comparing their fairness in terms of security and usability on two public data sets, involving six groups based on gender and ethnicity, showing that trends appearing in single-attribute analysis often disappear or reverse when multi-attribute groups are considered. 
\item We analyze the impact of data characteristics on models performance, through our framework, and observe a set of data characteristics with relevant explanatory power. 
\end{itemize}

Our results evidence the delicate entanglement between security, usability, fairness, and the characteristics of the image data received by the face recognition model.

\section{Related Work} \label{sec:related}

Our research is inspired by works in two areas that have direct impact on face recognition research: explorative unfairness analyses and unfairness countermeasures (Table~\ref{tab:prior-work}). 

\vspace{2mm} \noindent \textbf{Fairness Exploration in Face Biometrics}. 
Efforts were devoted to creating demographically balanced data sets and assessing performance differences among demographic and, rarely, non-demographic groups~\cite{inbook,9512390,9025435,demogpairs,terhorst2021comprehensive}. 
For instance, \cite{Ricanek2015ARO} showed that face recognition systems systematically provide children with worse performance than adults. 
Studies on gender highlighted that system's performance is on average lower for women than men~\cite{biasgenderbio,genderineq,sexist}. 
Regardless of the data imbalance, experiments with dimensionality reduction revealed that this disparate impact might appear since the similarity is higher between female than male faces. 
Disparities were also observed for under-represented ethnic groups~\cite{8638319,9186002}. 
Interestingly, model biases were similar to those present in humans' beliefs~\cite{humanslike}. 

Despite the increasing number of demographically balanced data sets, trained models may amplify biases as much as if data had not been balanced~\cite{balancednotenough}.
This finding motivated researchers to study whether other covariates associated to the face image (e.g., image distortions) or soft attributes (e.g., presence of beard) may generate disparities.
For instance, \cite{facequality} provided an analysis of the correlation between the quality of face images and bias estimates. 
Indeed, studies on attributes beyond demographics (e.g., the presence of glasses and the face pose) questioned that the reason behind poor performance is related to a protected attribute (and not to other covariates)~\cite{FairFace,terhorst2021comprehensive}.
Similarly, \cite{unravelling} analyzed whether distortions can bias model predictions.

\begin{figure*}[!t]
\centering
\includegraphics[width=.9\textwidth, trim=1 1 1 1,clip]{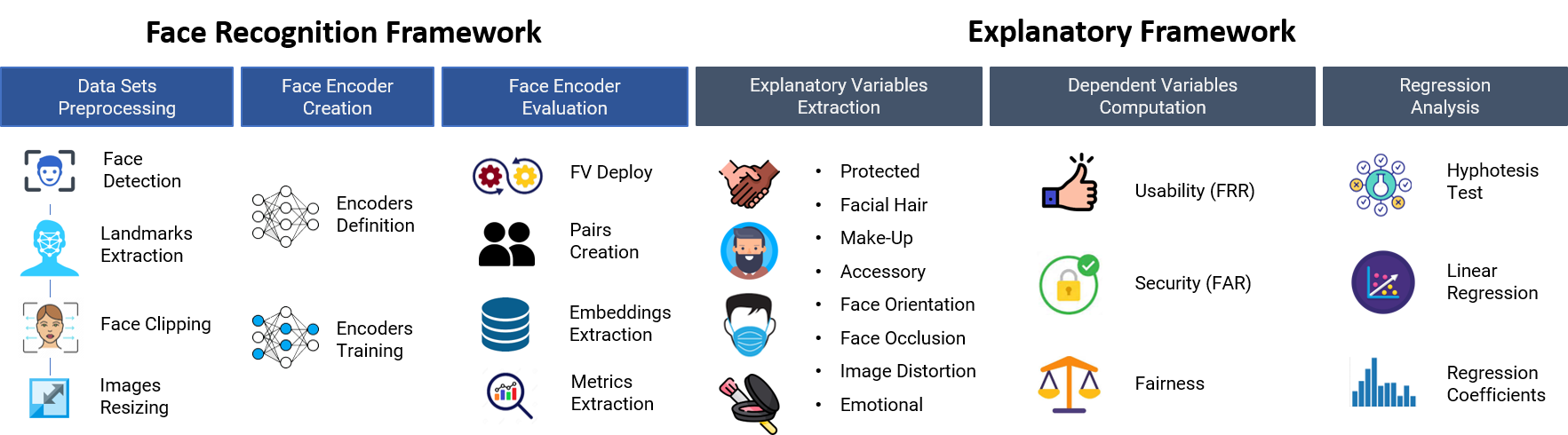}
\vspace{-3mm}
\caption{Once the face data sets were pre-processed, we created and evaluated a range of accurate face recognition models. We then ran an explanatory analysis on pre-trained models to assess the impact of data characteristics.}
\label{fig:pipeline}
\end{figure*}

\vspace{2mm} \noindent \textbf{Unfairness Mitigation in Face Biometrics}. 
Treatments against bias were applied in pre-processing, by modifying input data, in-processing, by constraining model training, and post-processing, by calibrating thresholds~\cite{2106.03761}.
Notable examples of in-processing treatments, e.g.,~\cite{kortylewski2019synthetic,popescu2021face}, consist of training models with synthetic images aimed to reduce imbalances across demographic groups.
Other works deliberately made the training data set imbalanced or filtered it to represent only one ethnic group at time~\cite{mitigate_racial}.
Considering in-processing treatments, \cite{comparisonlevelmitig} proposed a regularization term to minimize the mean absolute deviation in performance across groups.
Similarly, \cite{comparisonlevelmitig} defined two loss functions to reduce biased similarity estimates.
The method proposed in \cite{Wang2019RacialFI} combined supervised and unsupervised approaches to learn larger decision margins and reduce the discrimination. 
Moreover, multi-task learning involving accuracy and fairness was used in \cite{FairnessFaceMatch} to reduce bias. 
Finally, \cite{groupadaptivemitig} studied a loss function to narrow the intra-class distance between groups.
However, these treatments often showed a decrease in accuracy and an increase in computational requirements. 

\section{Problem Formulation}
We first formulate mathematically \emph{algorithmic discrimination} for face recognition. 
In our study, we focus on authentication, leaving identification as a future work. 

Formally, let $F \subset \mathbb{R}^{* \cdot * \cdot 3}$ denote the domain of images. 
We consider a feature extraction step which produces fixed-length representations in $D_\theta \subset \mathbb{R}^e$, under the feature extraction hyper-parameters $\theta$, where $e \in \mathbb N$. 
This step is formally denoted as $\mathcal{D}_{\theta}: F \rightarrow D_\theta$. 
Given a \textit{decision threshold} $\tau$, a verification trial can be defined as:

\vspace{-2mm}
\begin{equation} \label{eq:verification}
v_{\tau}: D_\theta \times D_\theta \rightarrow \{0,1\}    
\end{equation}

where an input feature vector $d_p \in D_\theta$ from an unknown user $p$ is compared with a feature vector $d_u \in D_\theta$ from user $u$ to confirm or refute the identity (1 and 0, respectively). 
Our protocol relies on a similarity function $\mathcal{S}: D \times D \rightarrow \mathbb{R}$: 

\vspace{-2mm}
\begin{equation} \label{eq:sim}
v_{\tau}(d_p \in D_\theta, d_u \in D_\theta) =  \mathcal{S}(d_p, d_u) > \tau
\end{equation}

Given individuals $\mathbb{I}$, training a face recognition model becomes an optimization problem aimed to maximize the expectation on the following objective function:

\vspace{-2mm}
\begin{equation} \label{eq:ex-func}
\underset{(\theta, \tau)}{\operatorname{argmax}} \mathop{\mathbb{E}}_{u, p \in \mathbb{I}} 
\begin{cases}
    v_{\tau} \left( d_p \in D_\theta, d_u \in D_\theta \right) & p = u \\
    1 - v_{\tau} \left( d_p \in D_\theta, d_u \in D_\theta \right) & p \neq u
  \end{cases} 
\end{equation}

We then define a face recognition system as being fair in case the probability $\mathbb{P}$ of an individual being (in)correctly authenticated is the same across demographic groups.  
Formally, given a ground-truth prediction $y=\{0, 1\}$ for a trial verification pair involving an individual from a certain group, assuming a binary case with a first group ($A=1$) and a second group ($A=0$), a face recognition system is considered as fair under our definition if it results in groups having equal probabilities for false positives ($v_{\tau}=1$ and $y=0$) and false negatives ($v_{\tau}=0$ and $y=1$), as follows:

\vspace{-2mm}
\begin{equation} \label{eq:equalodds}
    \setlength{\thickmuskip}{0mu}
    \scalebox{1.00}{$ \mathbb{P}(v_{\tau}|A=0,Y=y) = \mathbb{P}(v_{\tau}|A=1,Y =y)$}
\end{equation}

This definition is often referred to as \emph{equalized odds}, indicated also as conditional procedure accuracy equality. 

\section{Methodology}
Our methodology consists of two main phases (Figure \ref{fig:pipeline}).
We trained and calibrated a range of face recognition models.
Then, we leveraged the pre-trained models to run an exploratory analysis on the impact of data characteristics.

\subsection{Face Recognition Framework}
In this section, we show our experimental setup and the details of the used data sets, face encoders, and so on. 
We explain our model pre-training and threshold calibration, and provide an exhaustive evaluation of their performance. 
 
\begin{table*}[t!]
\label{table:3}
\begin{center}
\resizebox{\textwidth}{!}{%
\begin{tabular}{ll|rrr|rr|rr}
\hline
\textbf{Data Set Name}  & \textbf{Part}  & \textbf{\# Images} & \textbf{\# Users} & \textbf{Avg. Images/User} & \textbf{Avg. Images/Group} & \textbf{\# Groups}                   & \textbf{\% Minority}                   & \textbf{\% Majority}                       \\
\hline
DiveFace     & Training       & 111,503  & 19,196 & 6 $\pm$ 8           & 18,584 $\pm$ 2,968       & 6  & 49.07 (Male); 29.23 (Black)   & 50.92 (Female); 39.96 (Caucasian) \\
DiveFace     & Test       & 28,236  & 4,804 & 6 $\pm$ 11           & 4,706 $\pm$ 875       & 6  & 48.94 (Female); 27.72 (Black)   & 51.05 (Male); 39.91 (Caucasian) \\
VGG-Face2$^1$  & Test & 134,766 & 449  & 300 $\pm$ 108       & 22,461 $\pm$ 26,694         & 6  & 41.95 (Female); 4.71 (Black) & 58.04 (Male); 82.02 (Caucasian) \\ 
\hline
\multicolumn{8}{l}{$^1$ Our experiments did not include models trained on \texttt{VGG-Face2}. For training, we used only \texttt{DiveFace}, which ensures a good representation of all demographic groups.} \tabularnewline
\end{tabular}
}
\end{center}
\vspace{-6mm}
\caption{Relevant descriptive information of the two data sets considered in our study.}
\label{tab:data}
\end{table*}
\begin{table}[t!]
\begin{center}
\resizebox{1.00\linewidth}{!}{
\begin{tabular}{rr|rrr|rrr|r}
\hline
\multicolumn{2}{c}{\textbf{Face Models}$^{1, 2}$}                           & \multicolumn{3}{l}{\textbf{DiveFace}} & \multicolumn{3}{l}{\textbf{VGG-Face2}} & \multicolumn{1}{l}{}                                         \\
Backbone     & Head    & EER  & 1\%$^3$ & 0.1\%$^3$ & EER  & 1\%$^3$ & 0.1\%$^3$ & \#Params \\
\hline
ResNet152    & MagFace & 0.04 & 0.10      & 0.22        & 0.21 & 0.62      & 0.77        & 80M     \\
\ul{ResNet152}    & \ul{NPCFace} & \textbf{0.03} & 0.05      & 0.12        & \textbf{0.19} & 0.52      & 0.73        & 80M     \\
HRNet        & MagFace & 0.20 & 0.56      & 0.69        & 0.33 & 0.90      & 0.96        & 79M     \\
\ul{HRNet}        & \ul{NPCFace} & \textbf{0.05} & 0.11      & 0.24        & \textbf{0.21} & 0.60      & 0.81        & 79M     \\
AttentionNet & MagFace & 0.06 & 0.16      & 0.32        & 0.23 & 0.71      & 0.88        & 108M    \\
\ul{AttentionNet} & \ul{NPCFace} & \textbf{0.04} & 0.11      & 0.25        & \textbf{0.22} & 0.65      & 0.82        & 108M    \\
RepVGG       & MagFace & 0.07 & 0.28      & 0.44        & 0.28 & 0.80      & 0.93        & 116M    \\
\ul{RepVGG}       & \ul{NPCFace} & \textbf{0.07} & 0.18      & 0.34        & \textbf{0.28} & 0.74      & 0.91        & 116M    \\
ResNeSt      & MagFace & 0.10 & 0.85      & 0.88        & 0.26 & 0.77      & 0.88        & 86M     \\
\ul{ResNeSt}      & \ul{NPCFace} & \textbf{0.07} & 0.17      & 0.31        & \textbf{0.27} & 0.71      & 0.89        & 86M       \\
\hline
\multicolumn{9}{l}{$^1$ Due to space constraints, our study focused on the \ul{best setting} per backbone.} \tabularnewline
\multicolumn{9}{l}{$^2$ Models have an avg. size of 377.49$\pm$60 megabytes.} \tabularnewline
\multicolumn{9}{l}{$^3$ Reported values refer to FRRs w.r.t. the specified FAR percentage threshold.} \tabularnewline
\tabularnewline
\end{tabular}}
\end{center}
\vspace{-6mm}
\caption{Statistics on the considered face encoders.}
\label{tab:backbones}
\end{table}

\vspace{2mm} \noindent \textbf{Data Sets Preprocessing}. 
For our experiments, we chose \texttt{DiveFace}~\cite{diveface}, a key data set for fairness analysis in face recognition, and \texttt{VGGFace2}~\cite{cao2018vggface2}, a data set often used for face recognition benchmarking in general (see Table~\ref{tab:data}). 

\texttt{DiveFace} is composed of $140,000$ face images from $24,000$ identities, balanced across gender and ethnicity. 
Gender classes are binary, namely Woman and Man. Ethnicity classes include Asian, Black, and Caucasian. 
Training and test sets, including $70\%$ and $30\%$ of the identities respectively, were provided by the original authors. 
Their split ensured that demographic groups were equally represented in both training and test sets.
In contrast, \texttt{VGGFace2} consists of a training and a test set with $3.1$ million images and $170,000$ images respectively.
Our study considers only the test set, adopted as a validation for the findings observed in the first data set.
The gender and ethnicity classes were the same reported for \texttt{DiveFace}~\footnote{With DeepFace~\cite{serengil2020lightface}, we trained an ethnicity predictor on \texttt{DiveFace} and used it for inferring the ethnicity labels in the \texttt{VGGFace2} test set.}. 

The DeepFace~\cite{serengil2020lightface} toolkit was leveraged to (i) detect the face bounding box and the facial landmarks, (ii) clip, (ii) and resize the original images of both the data sets.

\vspace{2mm} \noindent \textbf{Face Encoder Creation}.
We built face encoders by combining CNN backbones (\texttt{ResNet152}, \texttt{AttentionNet}, \texttt{ResNeSt}, \texttt{RepVGG}, \texttt{HRNet}) and head classification networks (\texttt{MagFace}, \texttt{NPCFace}) proved to provide state-of-the-art performance in recent benchmarks~\cite{wang2021facex}.

In detail, \texttt{ResNet152} is a variant of the ResNet architecture~\cite{he2015deep} characterized by shortcut connections to obtain the residual counterpart.
\texttt{AttentionNet}~\cite{wang2017residual} is again a neural network with residuals, but enriched with attention modules. 
Each of these modules consists of (i) a mask branch acting as a gradient update filter during training and as a feature selector during inference and (ii) a trunk branch for feature processing during both phases.
\texttt{ResNeSt}~\cite{zhang2020resnest} is characterized by split-attention blocks, each including a feature map group (each characteristic can be divided into different groups based on a hyper-parameter) and split attention operations (a combined representation for each group can be obtained via an element-wise sum on multiples split).
\texttt{RepVGG}~\cite{ding2021repvgg} uses the relative simplicity of its structure as its strength. 
The inference is performed through a series of $3 \times 3$ and ReLU convolutions, while the training layers follow a multi-branch topology.
Unlike other backbones, \texttt{HRNet}~\cite{wang2020deep} maintains a high resolution along convolutions, thanks to parallel connections of convolutional streams and continuous exchange of information across resolutions.

Each backbone was plugged into a head network for the final classification.
\texttt{MagFace} is a refined implementation of the well-known ArcFace~\cite{deng2019arcface}. 
This head adds an additive angular margin penalty to enhance intra-class compactness and inter-class discrepancy.
\texttt{NPCFace}~\cite{zeng2020npcface} emphasizes the training on both negative and positive hard cases via the collaborative-margin mechanism in softmax logits. 

One face encoder per backbone - head network combination ($5 \times 2 = 10$ models) was optimized on the \texttt{DiveFace} training set\footnote{Being focused on demographic analysis, our models were trained on \texttt{DiveFace}, which shows a balanced representation of groups.}. 
Each face encoder was trained for 80 epochs at most (early stopping, patience $5$), with $64$-sized batches. The loss function was Categorical Cross-entropy, the optimizer was SGD, with momentum $0.9$, weight decay $1e-8$, initial learning rate $0.1$, and decays at $5, 25, 68$ epochs. 

\vspace{2mm} \noindent \textbf{Face Encoder Evaluation}.
Once trained, for each face encoder, we unplugged the classification head such that each face encoder would return as output the latent representation (size: $512$) of the face image given as input. 
With the face images of individuals included in each test set (disjoint set of individuals with respect to the training set), we then simulated a face verification task, by creating a range of trial verification pairs for each individual: 6 positive pairs\footnote{Since the minimum number of images per person was $4$, we could generate $(4 \times 3) / 2 = 6$ positive pairs from the images of each person.} with both images coming from the same person and 50 negative pairs with the second image in the pair coming from another person. 
Subsequently, for each face encoder and trial verification pair, we extracted the 512-sized latent representations of the two face images and then computed their Cosine similarity (range: $[-1, 1]$). 
Once we collected all the Cosine similarity scores resulting from a given face encoder, we determined three recognition thresholds respectively associated to the overall EER and the fixed False Acceptance Rates (FARs) of 1\% and 0.1\%.
The evaluation is summarized in Table~\ref{tab:backbones}.
Due to space constraints, our analysis in this paper focused on the EER threshold\footnote{Results on more secure thresholds had patterns similar in trend but more evident in magnitude with respect to those under the EER threshold.}. 

\subsection{Explanatory Framework}
In this section, we describe our explanatory framework aimed to investigate the impact of data characteristics on face recognition performance.
A core question in explanatory modeling research is the choice of explanatory variables. 
Two complementary approaches exist for choosing them, based on confirmatory or exploratory research. 
In confirmatory research, the potential impact of different variables are hypothesized a-priori, based on existing theories. 
This approach is useful when researchers have a theory supported by facts. 
The second exploration-driven approach is used when there exists a lack of sufficient theory foundations. 
Exploratory research could produce new hypotheses that could formally be evaluated later. 
Our framework is based on an exploration-driven approach.
With it, we studied the impact of data characteristics on recognition performance, in terms of false acceptance and rejection rates. 

\vspace{2mm} \noindent \textbf{Explanatory Variables Extraction}.
In addition to the gender and ethnicity labels already provided for the data sets, we extended the descriptive metadata accompanying each image with a range of attributes extracted via the \emph{Microsoft Cognitive Services}\footnote{\url{https://bit.ly/3PiVTvr}}. We specifically considered:

\begin{itemize}[leftmargin=10pt,itemsep=1pt]
\item \textbf{\ul{Protected attributes}} (shape: $3$) represent human characteristics it is against the law to discriminate someone because of. We included \emph{gender} (Man, Woman), \emph{ethnicity} (Asian, Black, Caucasian), and \emph{age} in $[1, 100]$. 
\item \textbf{Facial hair attributes} (shape: $3$) represent hair grown on the face. We considered the estimated presence of \emph{mustache} in $[0, 1]$, \emph{beard} in $[0, 1]$, and \emph{sideburns} in $[0, 1]$.  
\item \textbf{Make-up attributes} (shape: $2$) control the presence of cosmetics on the face. We considered the estimated presence of \emph{eye makeup} in $[0, 1]$ and \emph{lip makeup} in $[0, 1]$. 
\item \textbf{Accessory attributes} (shape: $2$) consider whether the person wears face accessories. We included the estimated presence of \emph{head wear} in $[0, 1]$ and \emph{glasses} in $[0, 1]$.  
\item \textbf{Face orientation attributes} (shape: $3$) describe the orientation of a face in a 3D space by the roll, yaw and pitch. We considered the order of three angles, \emph{roll}, \emph{yaw}, and \emph{pitch}, whose range is in $[-180,180]$ degrees. 
\item \textbf{Face occlusion attributes} (shape: $4$) show whether there are any parts of the face occluded. We included boolean flags for  \emph{occluded forehead}, \emph{occluded eyes}, \emph{occluded mouth}, and a continuous score for \emph{face exposure} in $[0, 1]$. 
\item \textbf{Image distortion attributes} (shape: $2$) monitor whether an image is unnaturally deformed. We included the estimated level of \emph{blur} in $[0, 1]$ and \emph{noise} in $[0, 1]$.
\item \textbf{Emotional attributes} (shape: $1$) indicate whether the person is conveying a certain emotion. We included the estimated presence of a \emph{smile} in $[0, 1]$. 
\end{itemize}

To support interpretation, the frequency or prevalence of values for each extracted variable on images from the two considered data sets is reported in the supplementary material.
Given an individual, we finally computed a 20-sized vector (one cell per variable), by computing the average or the mode of each variable across images of that individual.

\vspace{2mm} \noindent \textbf{Dependent Variables Computation}. The dependent variable is the performance of the face recognition model at user level as the false acceptance rate and the false rejection rate\footnote{We considered FAR and FRR separately, instead of the total error rate, to study if a group suffers from security or usability issues. Since we adopted the same EER threshold for all individuals, a FAR decrease for a group often resulted in a FRR increase for another group (and viceversa).}. 
To determine them, in our study, we considered the threshold $\tau_{EER}$ (same for all individuals and all demographic groups) that led to the EER based on the entire set of positive and negative trial verification pairs we computed.  

\textit{Usability}. The False Rejection Rate (FRR) represents the measure of the likelihood that the system will incorrectly reject an access attempt by an authorized individual. A system’s FRR is computed as the number of false rejections divided by the number of positive verification attempts. Given the verification pairs $\mathcal D_u$, the FRR for an individual $u$ is:

\vspace{-3mm}
\begin{equation} 
    FRR(u) = \frac{|\{ (u,p) \in \mathcal D_u | v_{\tau}(d_p, d_u) = 0 \cap u = p \}|}
                  {|\{ (u,p) \in \mathcal D_u | u = p \}|} 
\label{eq:frr}
\end{equation}

\textit{Security}. The False Acceptance Rate (FAR) is the measure of the likelihood that the system will incorrectly accept an access attempt by an unauthorized user. A system’s FAR is computed as the number of false acceptances divided by the number of negative verification attempts. Given the verification pairs $\mathcal D_u$, the FAR for an individual $u$ is:

\vspace{-2mm}
\begin{equation}
    FAR(u) = \frac{|\{ (u,p) \in \mathcal D_u | v_{\tau}(d_p, d_u) = 1 \cap u \neq p \}|}
                  {|\{ (u,p) \in \mathcal D_u | u \neq p \}|} 
\label{eq:far}
\end{equation}

Subsequently, the FAR (FRR) for a given demographic group was computed by averaging the individual FAR (FRR) of all members belonging to that group. 

\textit{Fairness}. We operationalize the equalized odds notion (Eq.~\ref{eq:equalodds}) through two evaluation metrics, which measure the difference in FAR (FRR) between groups. Formally, given two groups $g_i, g_j \subset \mathbb I$, with $g_i \cap g_j = \emptyset$, constructed by considering one or more protected attributes (e.g., gender only, ethnicity only, or their combination), these two evaluation metrics are defined as follows:

\vspace{-2mm}
\begin{equation} 
\Delta FRR(g_i, g_j) = FRR(g_i) - FRR(g_j)
\label{eq:fair-frr}
\vspace{-2mm}
\end{equation}

\vspace{-2mm}
\begin{equation}
\Delta FAR(g_i, g_j) = FAR(g_i) - FAR(g_j)
\label{eq:fair-far}
\end{equation}

A higher magnitude of $\Delta FAR$ and $\Delta FRR$ (either positive or negative) indicates a higher model unfairness. 

\vspace{2mm} \noindent \textbf{Regression Analysis}.
In explanatory modeling, we are interested in identifying variables that have a statistically significant relationship with a dependent variable. 
In our study, we analyzed whether some image characteristics (explanatory variables) can explain the variations of FAR or FRR (dependent variables). 
A multiple regression model was used to analyze these relationships, fitting it with a matrix of size ($|\mathbb{I}|, 20)$ representing the $20$ image characteristics per individual as an input, and a vector of size $|\mathbb{I}|$ representing the FAR (FRR) per individual under a given face encoder as an output. Formally, the model is defined as: 

\vspace{-2mm}
\begin{equation} \label{eq:lin-model}
   y = \epsilon + \gamma_0 + \sum_{j=1}^{f} \gamma_{j} \cdot c_{j}
\end{equation}

where $y$ is the dependent variable (either Eqs.~\ref{eq:frr} or \ref{eq:far}), and $\gamma_j$ and $c_{j}$ respectively represent the regression coefficient and the value of the $j$-th image characteristic.
The coefficients $\gamma \in \Gamma$ indicate us the extent to which a change of a given image characteristic, holding all the others constant, leads to a change in the FAR or FRR.
Being a multiple regression, our analysis tested the null hypothesis that all the regression coefficients $\gamma \in \Gamma$ are zero, versus the alternative that at least one coefficient $\gamma_j$ is nonzero.
We hence conjecture that image characteristics with a nonzero statistically significant coefficient might be deemed as important to explain disparate impacts among (groups of) individuals.

\section{Experimental Results}
Our experiments analyzed the extent to which unfairness occurs when we consider multiple protected attributes jointly (\emph{RQ1}), whether there is any co-relationship among image characteristics and error rates (\emph{RQ2}), and how a change in an image characteristic impacts error rates (\emph{RQ3}). 

\subsection{Multiple Protected Attribute Analysis (RQ1)}
In a first experiment, we investigated whether and how unfairness phenomena observed under a single protected attribute change when we consider multiple protected attributes simultaneously\footnote{To keep the study concise, since the age attribute is not categorical and we would avoid to use arbitrary thresholds to form age-based groups, we decided to consider groups based on gender and ethnicity only.}. Therefore, Table~\ref{tab:rq1} shows the average FARs and FRRs achieved by demographic groups created based on each protected attribute separately and based on a combination of protected attributes, for the five best performing models, on \texttt{DiveFace}. The results on \texttt{VGGFace2} and the statistical significance analyses for both data sets are provided in the supplementary material.

\begin{table}[!t]
\begin{subtable}{1.\linewidth}{
\resizebox{\textwidth}{!}{
\begin{tabular}{l|rra|rra}
\hline
{} & \multicolumn{3}{l}{\textbf{FAR}} & \multicolumn{3}{l}{\textbf{FRR}} \\
 &    \emph{Woman} &      \emph{Man} &       \emph{$W \cup M$} &    \emph{Woman} &      \emph{Man} &       \emph{$W \cup M$} \\
\hline
\emph{Asian}     &  0.048 &  0.043 &  0.046 &  0.045 &  0.042 &  0.043 \\
\emph{Black}     &  0.047 &  0.045 &  0.046 &  \textbf{0.050} &  0.041 &  0.045 \\
\emph{Caucasian} &  \textbf{0.051} &  0.049 &  0.050 &  0.048 &  0.036 &  0.042 \\
\rowcolor{boxgrey}
\emph{$A \cup B \cup C$}    &  0.049 &  0.045 &  0.047 &  0.048 &  0.040 &  0.044 \\
\hline
\end{tabular}}}
\vspace{-2mm}
\caption{AttentionNet + NPCFace}
\vspace{2mm}
\end{subtable}
\begin{subtable}{1.\linewidth}{
\resizebox{\textwidth}{!}{
\begin{tabular}{l|rra|rra}
\hline
{} & \multicolumn{3}{l}{\textbf{FAR}} & \multicolumn{3}{l}{\textbf{FRR}} \\
 &    \emph{Woman} &      \emph{Man} &       \emph{$W \cup M$} &    \emph{Woman} &      \emph{Man} &       \emph{$W \cup M$} \\
\hline
\emph{Asian}     &  \textbf{0.055} &  0.048 &  0.051 &  0.041 &  0.043 &  0.042 \\
\emph{Black}     &  0.044 &  0.041 &  0.043 &  \textbf{0.056} &  0.048 &  0.052 \\
\emph{Caucasian} &  0.053 &  0.051 &  0.052 &  0.047 &  0.041 &  0.044 \\
\rowcolor{boxgrey}
\emph{$A \cup B \cup C$}      &  0.051 &  0.046 &  0.049 &  0.048 &  0.044 &  0.046 \\
\hline
\end{tabular}}}
\vspace{-2mm}
\caption{HRNet + NPCFace}
\vspace{2mm}
\end{subtable}
\begin{subtable}{1.\linewidth}{
\resizebox{\textwidth}{!}{
\begin{tabular}{l|rra|rra}
\hline
{} & \multicolumn{3}{l}{\textbf{FAR}} & \multicolumn{3}{l}{\textbf{FRR}} \\
 &    \emph{Woman} &      \emph{Man} &       \emph{$W \cup M$} &    \emph{Woman} &      \emph{Man} &       \emph{$W \cup M$} \\
\hline
\emph{Asian}     &  \textbf{0.080} &  0.067 &  0.073 &  0.067 &  0.063 &  0.065 \\
\emph{Black}     &  0.065 &  0.057 &  0.061 &  0.084 &  \textbf{0.087} &  0.085 \\
\emph{Caucasian} &  0.079 &  0.077 &  0.078 &  0.061 &  0.055 &  0.058 \\
\rowcolor{boxgrey}
\emph{$A \cup B \cup C$}     &  0.075 &  0.067 &  0.071 &  0.071 &  0.068 &  0.069 \\
\hline
\end{tabular}}}
\vspace{-2mm}
\caption{RepVGG + NPCFace}
\vspace{2mm}
\end{subtable}
\begin{subtable}{1.\linewidth}{
\resizebox{\textwidth}{!}{
\begin{tabular}{l|rra|rra}
\hline
{} & \multicolumn{3}{l}{\textbf{FAR}} & \multicolumn{3}{l}{\textbf{FRR}} \\
 &    \emph{Woman} &      \emph{Man} &       \emph{$W \cup M$} &    \emph{Woman} &      \emph{Man} &       \emph{$W \cup M$} \\
\hline
\emph{Asian}     &  \textbf{0.077} &  0.068 &  0.072 &  0.058 &  0.069 &  0.064 \\
\emph{Black}     &  0.057 &  0.054 &  0.055 &  0.064 &  0.071 &  0.065 \\
\emph{Caucasian} &  0.076 &  0.069 &  0.073 &  \textbf{0.078} &  0.055 &  0.067 \\
\rowcolor{boxgrey}
\emph{$A \cup B \cup C$}    &  0.070 &  0.064 &  0.067 &  0.067 &  0.065 &  0.066 \\
\hline
\end{tabular}}}
\vspace{-2mm}
\caption{ResNeSt + NPCFace}
\vspace{-2mm}
\end{subtable}
\begin{subtable}{1.\linewidth}{
\vspace{2mm}
\resizebox{\textwidth}{!}{
\begin{tabular}{l|rra|rra}
\hline
{} & \multicolumn{3}{l}{\textbf{FAR}} & \multicolumn{3}{l}{\textbf{FRR}} \\
 &    \emph{Woman} &      \emph{Man} &       \emph{$W \cup M$} &    \emph{Woman} &      \emph{Man} &       \emph{$W \cup M$} \\
\hline
\emph{Asian}     &  \textbf{0.034} &  0.029 &  0.032 &  0.024 &  0.029 &  0.026 \\
\emph{Black}     &  0.028 &  0.022 &  0.025 &  \textbf{0.035} &  0.027 &  0.031 \\
\emph{Caucasian} &  0.034 &  0.029 &  0.032 &  0.028 &  0.021 &  0.025 \\
\rowcolor{boxgrey}
\emph{$A \cup B \cup C$}      &  0.032 &  0.027 &  0.029 &  0.029 &  0.026 &  0.027 \\
\hline
\end{tabular}}}
\vspace{-2mm}
\caption{ResNet152 + NPCFace}
\vspace{2mm}
\end{subtable}
\vspace{-3mm}
\caption{Protected attribute analysis - \texttt{DiveFace}.}
\label{tab:rq1}
\end{table}   

From a security perspective (FAR, left-most columns in Table~\ref{tab:rq1}), it can be observed that the $\Delta$ FAR between gender groups was between $0.004$ (AttentionNet + NPCFace) and $0.008$ (RepVGG + NPCFace), and that all models resulted in a higher FAR for women than men (see $A \cup B \cup C$ rows). \emph{If we had looked only at the gender attribute, we would have concluded that the considered models systematically put women more at risk than men}. Considering ethnicity groups, Caucasian individuals experienced the highest FAR consistently across models, followed by Asian and then Black individuals (see $W \cup M$ columns). The exception was represented by Asian and Black groups with AttentionNet + NPCFace, where we measured the same FAR estimate for both groups. Similarly to the gender groups, \emph{if we had looked only at ethnicity, we would have concluded that Caucasian individuals were systematically disadvantaged}. Combining both protected attributes, these two conclusions would not still be necessarily valid. For instance, in HRNet + NPCFace, Caucasian men ($0.051$) resulted in a higher FAR than Black women ($0.044$)\footnote{These examples had a statistically significant difference ($p < 0.1$).}. This would collide with our conclusion on the gender attribute only. In RepVGG + NPCFace, the FAR for Asian women ($0.080$) and Caucasian women ($0.079$) was statistically indistinguishable, rubbing up against our conclusion on the ethnicity attribute only. 

\begin{figure*}[!t]%
    \centering
    \hspace{0.8cm}
    \subfloat[Dependent variable: FAR]{
    \includegraphics[width=.7\linewidth, trim=4 4 4 4,clip]{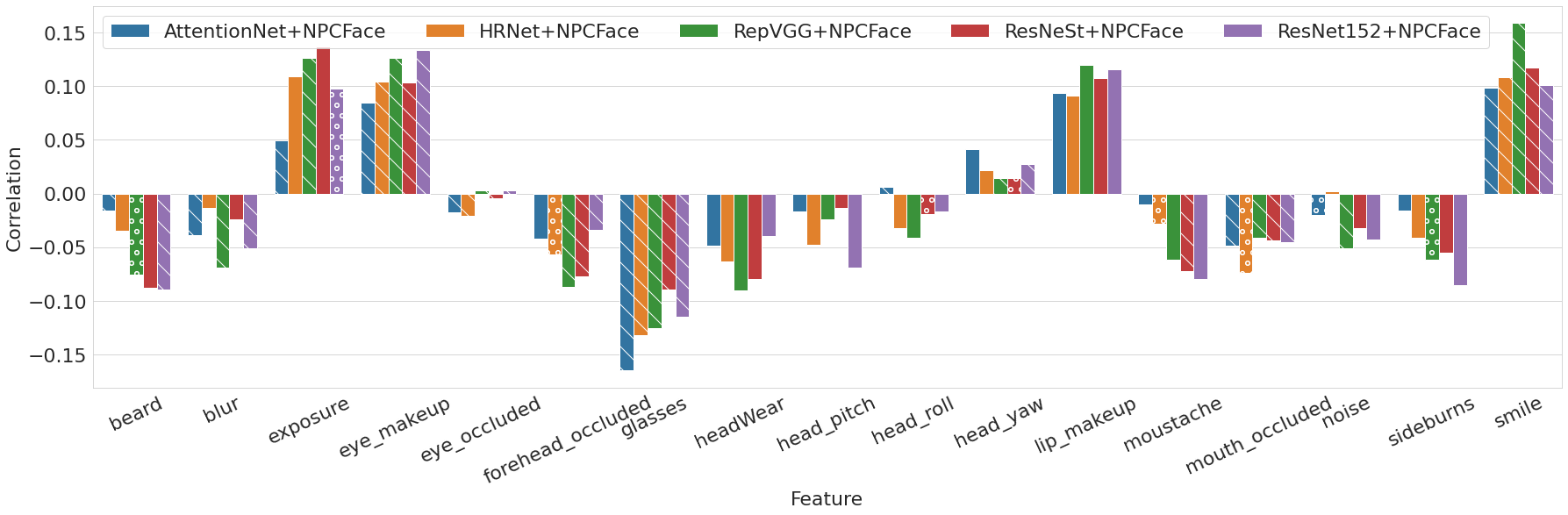}}
    \newline
    \subfloat[Dependent variable: FRR]{
    \includegraphics[width=.7\linewidth, trim=4 4 4 4,clip]{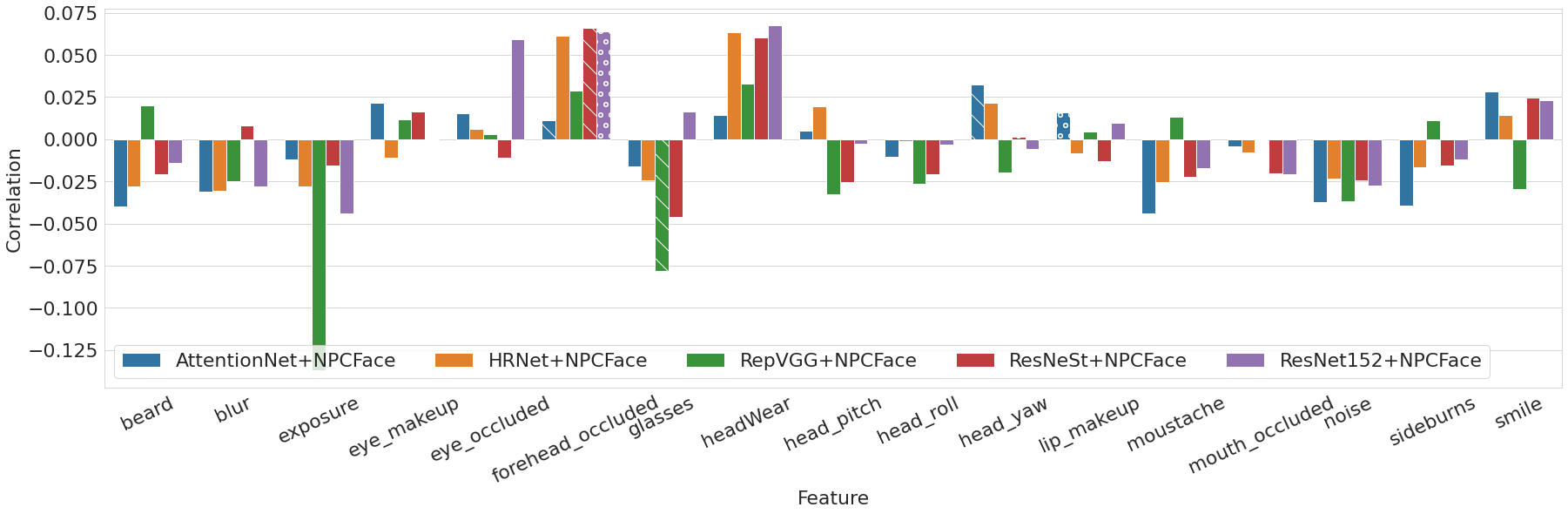}}
    \vspace{-2mm}
    \caption{Pearson correlation between an image characteristic for an individual and their corresponding estimated error rate, under the EER threshold (same for all individuals), on \texttt{DiveFace}. Significance level: (o) $p < 0.05$;  (\textbackslash) $p < 0.01$. \label{fig:rq2-corr}}
\end{figure*}

The relative ranking of demographic groups from a usability perspective (FRR, right-most columns in Table~\ref{tab:rq1}) varied more than from a security perspective. Specifically, women were disadvantaged the most, though the differences in FRR between gender groups were lower or even negligible. Again, \emph{if we had looked only at the gender attribute, we would have concluded that the considered models were less usable for women than men}. Considering ethnicity, most of the models were less fair to Black individuals, while the same models varied more in their treatment of Caucasian and Asian. Specifically, HRNet + NPCFace and ResNeSt + NPCFace resulted in a higher FRR for Caucasian than Asian individuals, viceversa for the other models. \emph{If we had looked only at ethnicity, we would have concluded again that Caucasian individuals were systematically disadvantaged}. Considering fine-grained groups, these conclusions often change. For instance, the RepVGG + NPCFace achieved a higher FRR for Black men ($0.087$) then Caucasian women ($0.061$), reversing both the gender attribute only conclusion and the ethnicity only conclusion.   

\hlbox{RQ1 Summary}{Trends appearing in demographic groups under a single protected attribute often disappear or even reverse when these groups are framed more, from both security and usability perspectives. } 

\subsection{Correlation Coefficients Analysis (RQ2)}
In a second analysis, we were interested in determining whether any co-relationship or association exists between the considered (non-protected) image characteristics and the error rates experienced by a face recognition in terms of security (FAR) and usability (FRR).
Indeed, the findings uncovered in RQ1 showed that unfairness issues are complex and may go beyond a mere membership to a certain demographic group. We hence conducted a correlation analysis between image characteristics relevant in prior work, e.g., \cite{terhorst2021comprehensive}, and each dependent variable (FAR or FRR)\footnote{Though two variables X and Y may seem to be linked, it might be possible but not certain that X caused Y, since it could be also that Y caused X or that a third variable Z caused both X and Y. For instance, the main cause of women-men disparities might be the gender attribute itself, and the presence of facial hair and cosmetics might be the secondary dependent factor. Conversely, the presence of facial hair and cosmetics might be the main cause and the gender attribute itself might be the secondary factor (different proportions of women and men having facial hair or cosmetics).}. 

Figure~\ref{fig:rq2-corr} shows correlation coefficients between the image characteristics and the dependent variables for the five best performing models on \texttt{DiveFace}). The correlation coefficients on \texttt{VGGFace2} are provided in the supplementary material.  
FARs were positively (negatively) correlated with non-protected attributes like make-up, exposure, and smile (facial hair, face occlusion and orientation, and image distortions). FRRs were positively (negatively) correlated with face occlusion and orientation and smile (facial hair, exposure, accessories, and face occlusion). 
Correlation coefficients for the same attribute varied across models, confirming the complex relationships among image characteristics, models, and disparate impacts. 

\hlbox{RQ2 Summary}{There exist co-relationships between error rates (FAR and FRR) and non-protected attributes, especially make-up and facial hair.}
\vspace{1mm}

\subsection{Regression Coefficients Analysis (RQ3)}
In a third analysis, motivated by the co-relationships uncovered in RQ2, we leveraged our explanatory framework to estimate the extent to which a change in an image characteristic impacts the estimated FAR and FRR (dependent variables). 
To this end, we fitted a multiple regression model with the extracted image characteristics per individual as an input and the FAR (FRR) per individual under a face encoder on \texttt{DiveFace} as an output. 
The results on \texttt{VGGFace2} are reported in the supplementary material. 

Figure~\ref{fig:rq2-2} shows the regression coefficients and their significance for the five best performing models, respectively considering FAR and FRR as dependent variables. 
It can be observed that a change in smile and glasses seemed to significantly impact FAR, regardless of the face encoder.
Exposure, head roll, lip makeup, and mouth occluded were significant only on few face encoders. 
Though some of the other image characteristics had a nonzero coefficient, they were not considered significant.
On the other hand, the statistically significant image characteristics for FAR were less than for FAR (exposure, eye occluded, glasses, head wear).

\hlbox{RQ3 Summary}{Changes on certain non-protected attributes significantly impact error rates, especially FAR, regardless of the face encoder.} 

\section{Discussion and Conclusion}
In this section, we connect the insights coming from individual analysis and discuss the results and future research.

\begin{figure}[!t]
\centering
\subfloat[Dependent variable: FAR\label{fig:fine40-tuned}]{
   \includegraphics[width=1.\linewidth, trim=4 4 4 4,clip]{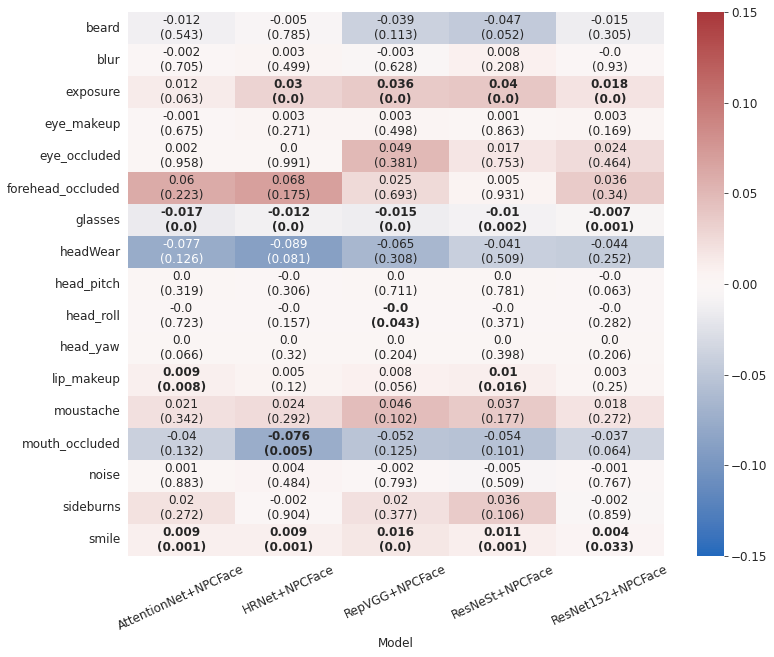}}
 \newline
\subfloat[Dependent variable: FRR\label{fig:fine40-tuned}]{
   \includegraphics[width=1.\linewidth, trim=4 4 4 4,clip]{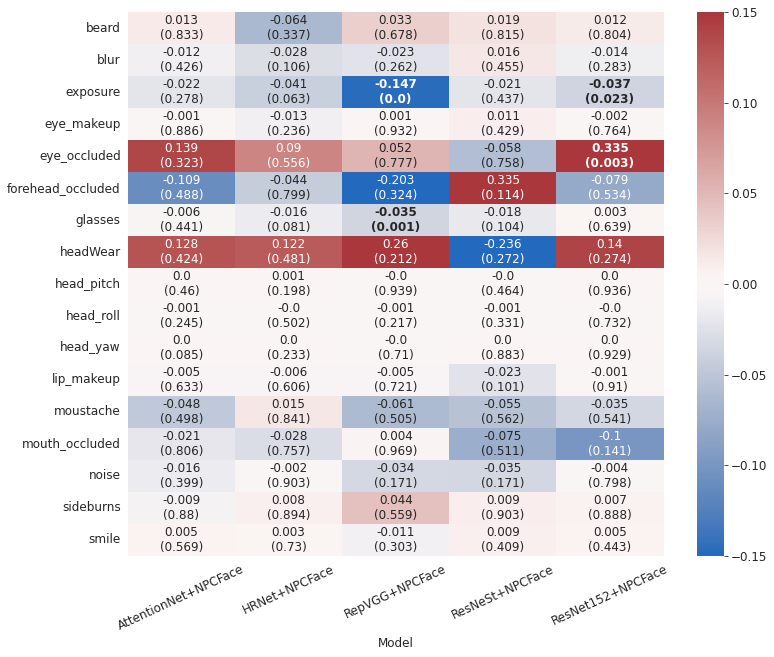}}
\caption{Regression coefficient (p-value) per characteristic with FAR and FRR as dependent variables, on \texttt{DiveFace}.}
\label{fig:rq2-2}
\end{figure}

Our first analysis (RQ1) revealed that trends appearing in groups created by considering only one protected attribute often disappear or even reverse when the groups are framed more. 
This uncovers the need of a more fine-grained analysis of performance across demographic groups.
In a second analysis (RQ2), our results highlighted that error rates have co-relationships with other covariates than mere protected attributes, especially make-up and facial hair. 
The effects of such covariates could be mitigated by, for instance, introducing specific pre-processing phases during data acquisition. 
Finally, our results (RQ3) confirmed that changes on certain non-protected attributes significantly impact error rates, especially FAR, regardless of the face encoder. 
Therefore, not only some key co-relationships between error rates and image characteristics exist, but also they might play a key role on the system's security and usability. 

As next steps, since the image characteristics extraction was based on an opaque black-box, we plan to devise more transparent models for their computation.
We will also extend the set of considered characteristics to better understand the underlying phenomena. 
Furthermore, contrasting observations resulted according to the data set, possibly also due to their different nature and the use of same training set for all the models.
We hence plan to conduct a more extensive model analysis, by extending (the combination of) training-test data sets from the literature and including an assessment of recognition performance under more secure thresholds.  
Our regression analysis depends on the number of image characteristics and their granularity. 
Its results could hence change depending on the number of protected attributes the fairness metric aims to monitor. 
Nonetheless, our framework could be applied independently of the number of attribute classes.
We will also analyze dependencies via multi-factor tests and image-level explainability techniques.
Finally, we will study mitigation treatments not requiring explicit knowledge on protected attributes. 

\balance
{\small
\bibliographystyle{ieee}
\bibliography{egbib}
}

\appendix

\onecolumn

{
    \centering
    {\large  \textbf{Supplementary Material for Manuscript \\ "Explaining Bias in Deep Face Recognition via Image Characteristics"}} \\
    \vspace{1mm} Andrea Atzori, Gianni Fenu, Mirko Marras \\
}

\counterwithin{figure}{section}
\counterwithin{table}{section}

\vspace{5mm} \noindent The provided additional material includes:

\begin{itemize}
  \setlength\itemsep{0em}
\item The distribution plots for the scores of each individual explanatory variable under both \texttt{DiveFace} and \texttt{VGGFace2}. 
\item The averaged FAR and FRR achieved by groups created based on single/multiple protected attributes on \texttt{VGGFace2}.
\item The outcomes of Kruskal-Willis statistical tests ran on FAR and FRR distribution of different groups on both data sets.
\item The correlation coefficients between image characteristics and the FAR (FRR) dependent variables on \texttt{VGGFace2}. 
\item The regression coefficients and their statistical significance for models fitted on the \texttt{VGGFace2} data set. 
\end{itemize}

\section{Explanatory Variables Distribution}
\begin{figure*}[!b]
    \centering
    \includegraphics[width=0.95\linewidth]{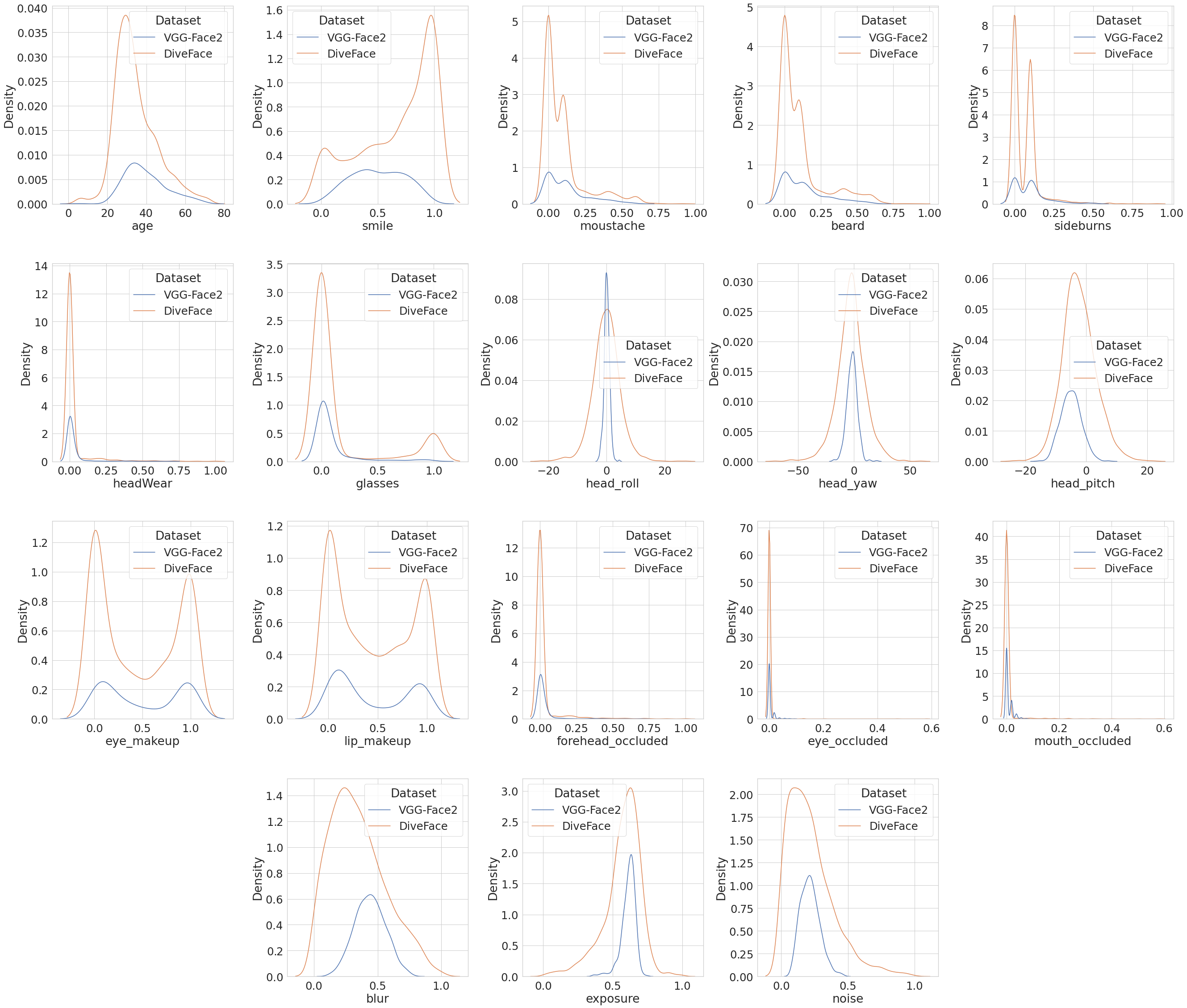}
    \caption{Explanatory variables distribution across data sets.}
    \label{fig:attr_dist}
\end{figure*}

To have a more detailed picture of the explanatory variable scores extracted via the \emph{Microsoft Cognitive Services}, Figure \ref{fig:attr_dist} depicts the distribution of these scores under both \texttt{DiveFace} and \texttt{VGGFace2} data sets. 
 
\section{Extended Results on Multiple Protected Attribute Analysis (RQ1)}
In a first experiment, we investigated whether and how unfairness phenomena observed under a single protected attribute change when we consider multiple protected attributes jointly. To complement the results on \texttt{DiveFace}, Table~\ref{tab:rq1-vgg} shows averaged FARs and FRRs achieved by groups created based on each protected attribute separately and based on a combination of protected attributes, for the five best performing models, on \texttt{VGGFace2}.

It can be observed that the error rates on this data set were generally higher than those measured on \texttt{DiveFace}. 
Similarly, the differences in averaged FAR and FRR were larger among demographic groups. 
Nevertheless, again several patterns observed under a single protected attribute ended up disappearing or even being reversed under multiple protected attributes. 
Considering the FRR obtained with the ResNet152 + NPCFace model, Caucasians had on average a lower FRR than Asians. 
This was no longer necessarily true on fine-grained groups, since Asian men and Caucasian women both experienced a FRR of $0.133$.
Similar observations can be made for the FAR of Asians and Blacks on RepVGG + NPCFace, as another example. 

\begin{table}[!b]
\vspace{2mm}
\centering
\begin{subtable}{.62\textwidth}{
\resizebox{\textwidth}{!}{
\begin{tabular}{l|rra|rra}
\hline
{} & \multicolumn{3}{l}{\textbf{FAR}} & \multicolumn{3}{l}{\textbf{FRR}} \\
 &    \emph{Woman} &      \emph{Man} &       \emph{$W \cup M$} &    \emph{Woman} &      \emph{Man} &       \emph{$W \cup M$} \\
\hline
\emph{Asian}     &  0.293 &  0.117 &  0.205 &  \textbf{0.300} &  0.200 &  0.250 \\
\emph{Black}     &  0.180 &  0.172 &  0.176 &  0.200 &  0.200 &  0.200 \\
\emph{Caucasian} &  \textbf{0.330} &  0.224 &  0.277 &  0.133 &  0.200 &  0.167 \\
\rowcolor{boxgrey}
\emph{$A \cup B \cup C$}    &  0.268 &  0.171 &  0.219 &  0.211 &  0.2 &  0.206 \\
\hline
\end{tabular}}}
\vspace{-2mm}
\caption{AttentionNet + NPCFace}
\vspace{2mm}
\end{subtable}
\begin{subtable}{.62\textwidth}{
\resizebox{\textwidth}{!}{
\begin{tabular}{l|rra|rra}
\hline
{} & \multicolumn{3}{l}{\textbf{FAR}} & \multicolumn{3}{l}{\textbf{FRR}} \\
 &    \emph{Woman} &      \emph{Man} &       \emph{$W \cup M$} &    \emph{Woman} &      \emph{Man} &       \emph{$W \cup M$} \\
\hline
\emph{Asian}     &  0.320 &  0.130 &  0.225 &  \textbf{0.367} &  0.133 &  0.250 \\
\emph{Black}     &  0.172 &  0.162 &  0.167 &  0.240 &  0.275 &  0.258 \\
\emph{Caucasian} &  \textbf{0.333} &  0.252 &  0.293 &  0.133 &  0.120 &  0.127 \\
\rowcolor{boxgrey}
\emph{$A \cup B \cup C$}      &  0.275 &  0.182 &  0.228 &  0.247 &  0.176 &  0.211 \\
\hline
\end{tabular}}}
\vspace{-2mm}
\caption{HRNet + NPCFace}
\vspace{2mm}
\end{subtable}
\begin{subtable}{.62\textwidth}{
\resizebox{\textwidth}{!}{
\begin{tabular}{l|rra|rra}
\hline
{} & \multicolumn{3}{l}{\textbf{FAR}} & \multicolumn{3}{l}{\textbf{FRR}} \\
 &    \emph{Woman} &      \emph{Man} &       \emph{$W \cup M$} &    \emph{Woman} &      \emph{Man} &       \emph{$W \cup M$} \\
\hline
\emph{Asian}     &  0.333 &  0.163 &  0.248 &  \textbf{0.367} &  0.133 &  0.250 \\
\emph{Black}     &  0.228 &  0.205 &  0.216 &  0.280 &  0.225 &  0.252 \\
\emph{Caucasian} &  \textbf{0.417} &  0.284 &  0.350 &  0.167 &  0.360 &  0.263 \\
\rowcolor{boxgrey}
\emph{$A \cup B \cup C$}     &  0.326 &  0.217 &  0.272 &  0.271 &  0.239 &  0.255 \\
\hline
\end{tabular}}}
\vspace{-2mm}
\caption{RepVGG + NPCFace}
\vspace{2mm}
\end{subtable}
\begin{subtable}{.62\textwidth}{
\resizebox{\textwidth}{!}{
\begin{tabular}{l|rra|rra}
\hline
{} & \multicolumn{3}{l}{\textbf{FAR}} & \multicolumn{3}{l}{\textbf{FRR}} \\
 &    \emph{Woman} &      \emph{Man} &       \emph{$W \cup M$} &    \emph{Woman} &      \emph{Man} &       \emph{$W \cup M$} \\
\hline
\emph{Asian}     &  0.336 &  0.153 &  0.245 &  0.333 &  0.233 &  0.283 \\
\emph{Black}     &  0.260 &  0.195 &  0.227 &  \textbf{0.360} &  0.250 &  0.305 \\
\emph{Caucasian} &  \textbf{0.413} &  0.252 &  0.332 &  0.133 &  0.240 &  0.186 \\
\rowcolor{boxgrey}
\emph{$A \cup B \cup C$}    &  0.336 &  0.200 &  0.268 &  0.275 &  0.241 &  0.258 \\
\hline
\end{tabular}}}
\vspace{-2mm}
\caption{ResNeSt + NPCFace}
\vspace{-2mm}
\end{subtable}
\begin{subtable}{.62\textwidth}{
\vspace{2mm}
\resizebox{\textwidth}{!}{
\begin{tabular}{l|rra|rra}
\hline
{} & \multicolumn{3}{l}{\textbf{FAR}} & \multicolumn{3}{l}{\textbf{FRR}} \\
 &    \emph{Woman} &      \emph{Man} &       \emph{$W \cup M$} &    \emph{Woman} &      \emph{Man} &       \emph{$W \cup M$} \\
\hline
\emph{Asian}     &  \textbf{0.227} &  0.163 &  0.195 &  0.233 &  0.133 &  0.183 \\
\emph{Black}     &  0.128 &  0.138 &  0.133 &  \textbf{0.240} &  0.175 &  0.208 \\
\emph{Caucasian} &  0.237 &  0.248 &  0.242 &  0.133 &  0.080 &  0.107 \\
\rowcolor{boxgrey}
\emph{$A \cup B \cup C$}      &  0.197 &  0.183 &  0.190 &  0.202 &  0.129 &  0.166 \\
\hline
\end{tabular}}}
\vspace{-2mm}
\caption{ResNet152 + NPCFace}
\vspace{2mm}
\end{subtable}
\caption{Protected attribute analysis - \texttt{VGGFace2}.\label{tab:rq1-vgg}}
\end{table}

\begin{figure*}[!b]
    \centering
    \subfloat[\centering DiveFace]{{\includegraphics[width=.3975\textwidth]{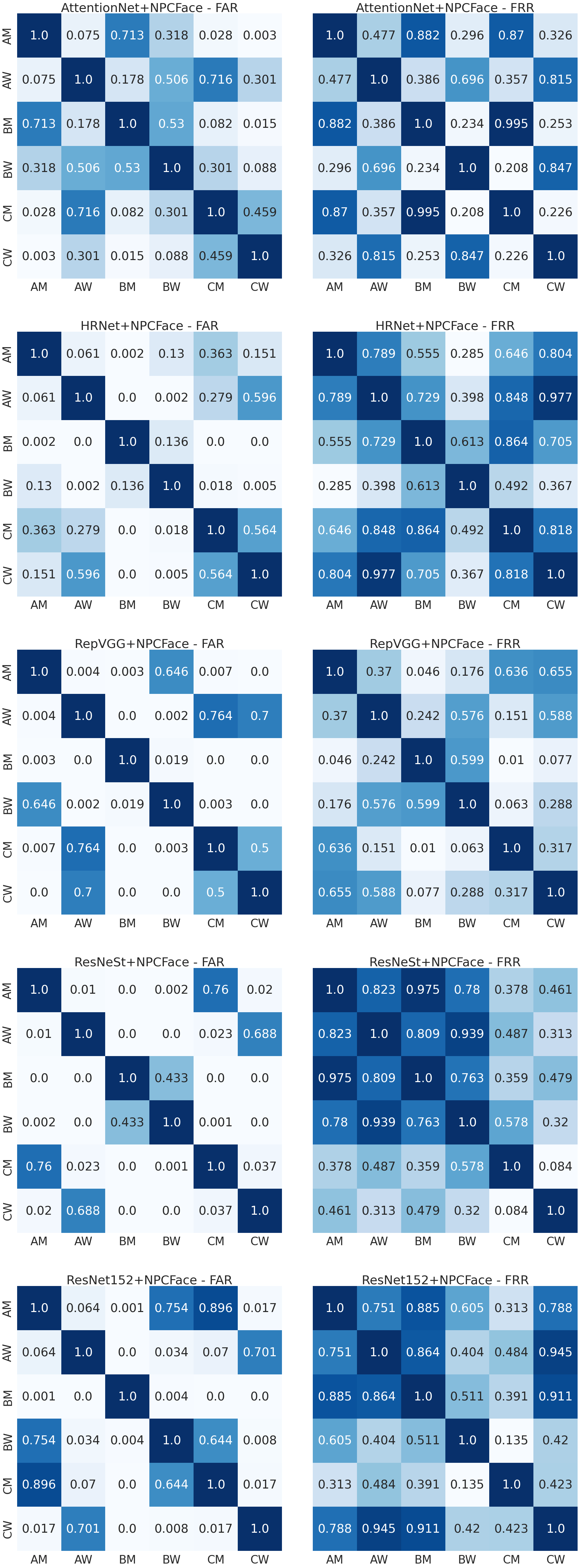}}}%
    \hspace{1cm}%
    \subfloat[\centering VGGFace2]{{\includegraphics[width=.3975\textwidth]{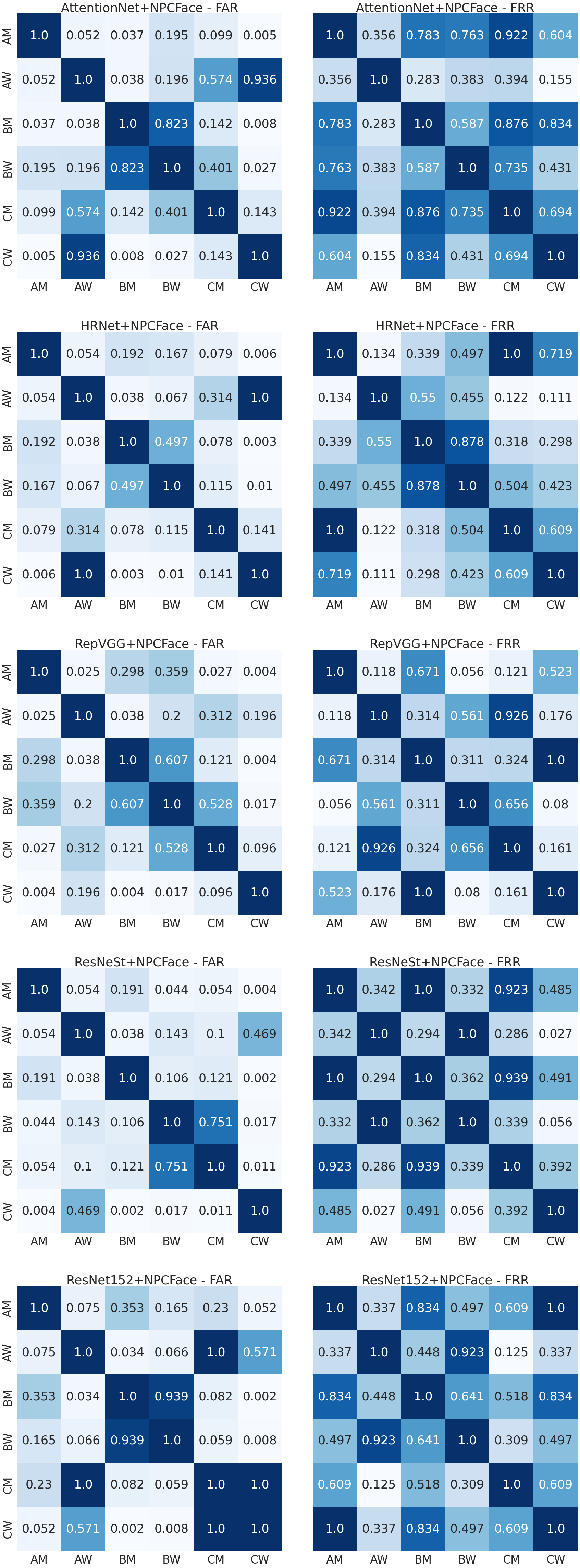}}}%
    \caption{P-values from a Kruskal-Willis statistical test between FAR and FRR distributions of demographic groups.}%
    \label{fig:distribution_similarities}%
\end{figure*}

Furthermore, we investigated whether the observed differences in FAR and FRR between groups were statistically significant. 
To this end, we ran a pair-wise Kruskal-Willis statistical tests for the FAR and FRR distributions of demographic groups, testing whether such distributions were equal or not. 
Specifically, the null hypothesis was that the distributions of all samples are equal.
Figure \ref{fig:distribution_similarities} collects the p-values resulting from the mentioned statistical tests. 

\section{Extended Results on Correlation Coefficients Analysis (RQ2)}
\begin{figure*}[!t]%
    \centering
    \hspace{0.8cm}
    \subfloat[Dependent variable: FAR]{
    \includegraphics[width=.99\linewidth, trim=4 4 4 4,clip]{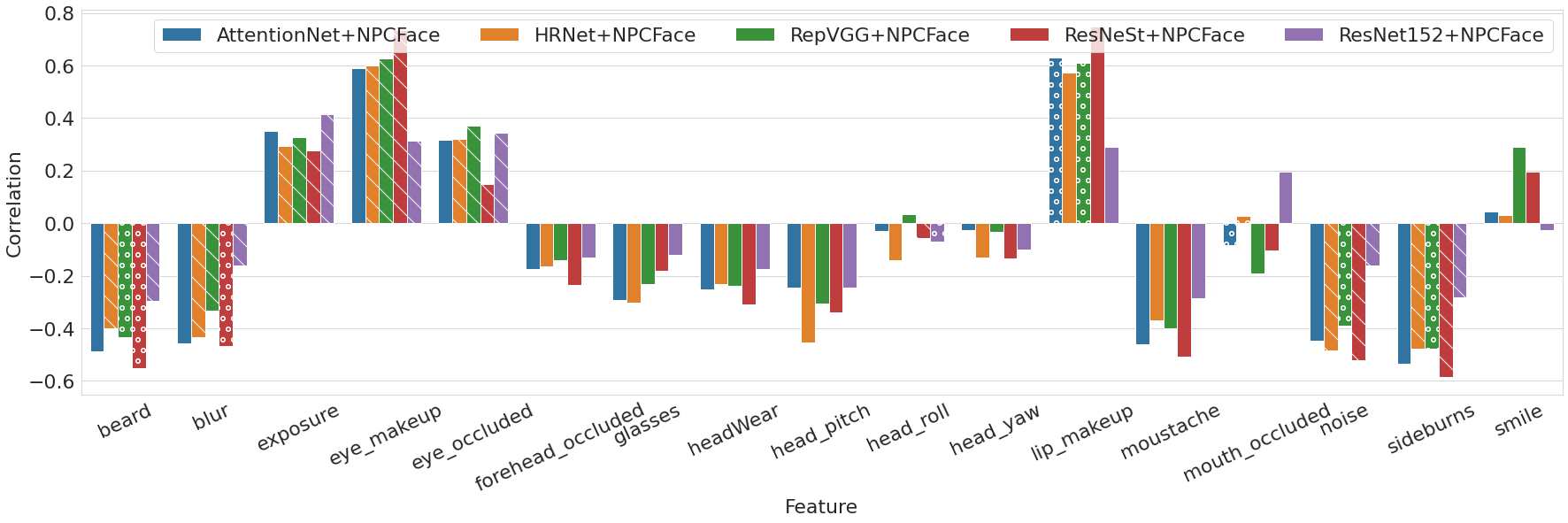}}
    \newline
    \newline
    \newline
    \subfloat[Dependent variable: FRR]{
    \includegraphics[width=.99\linewidth, trim=4 4 4 4,clip]{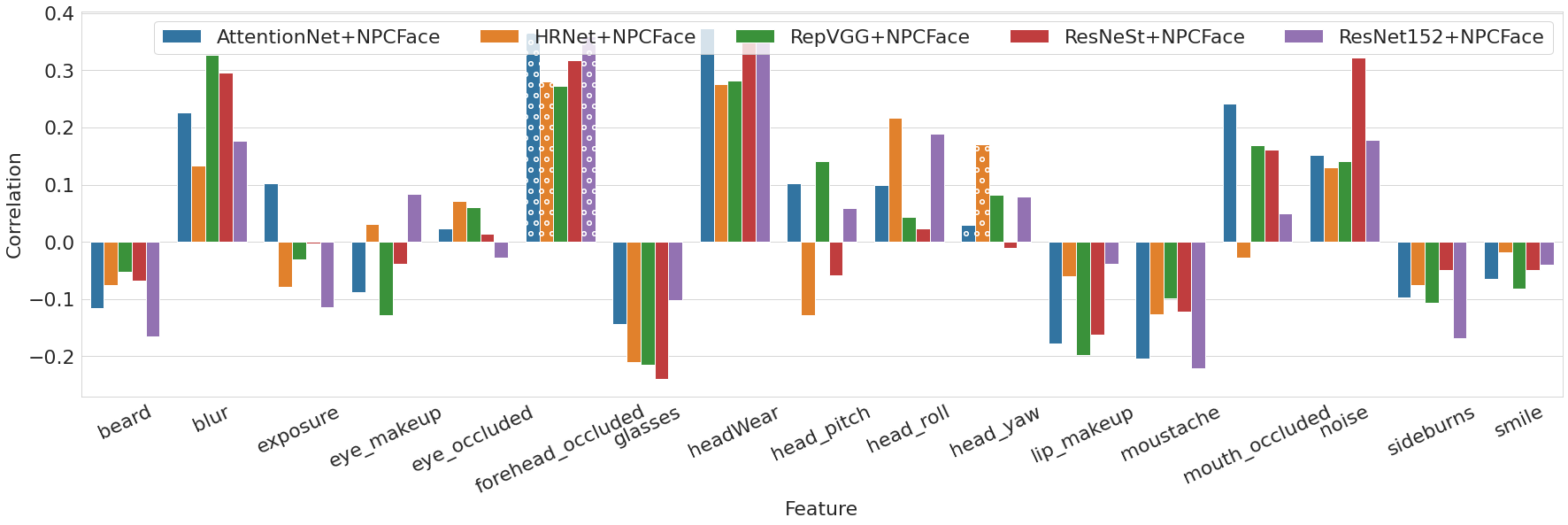}}
    \caption{Pearson correlation between a face characteristic estimate for images of an individual and their corresponding error rate under the EER threshold on \texttt{VGGFace2}. Significance level: (o) $p < 0.05$;  (\textbackslash) $p < 0.01$. \label{fig:rq2-corrvgg}}
    \vspace{15mm}
\end{figure*}

In a second analysis, we were interested in determining whether any co-relationship or association exists between the considered (non-protected) image characteristics and the error rates experienced by a face recognition in terms of security (FAR) and usability (FRR).
To this end, Figure~\ref{fig:rq2-corrvgg} shows correlation coefficients between the image characteristics and the dependent variables for the best five performing models on \texttt{VGGFace2}. 

\section{Extended Results on Regression Coefficients Analysis (RQ3)}
In a third analysis, motivated by the co-relationships uncovered in RQ2, we leveraged our explanatory framework to estimate the extent to which a change in an image characteristic (independent variable) impacts the estimated FAR and FRR (dependent variable). 
To this end, we fitted a multiple regression model with the extracted image characteristics per individual as an input and the FAR (FRR) per individual under a given face encoder on \texttt{VGGFace2} as an output. 

\begin{figure*}[!t]
\centering
\hspace{1.2cm}
\subfloat[Dependent variable: FAR\label{fig:fine40-tuned}]{
   \includegraphics[width=.65\linewidth, trim=4 4 4 4,clip]{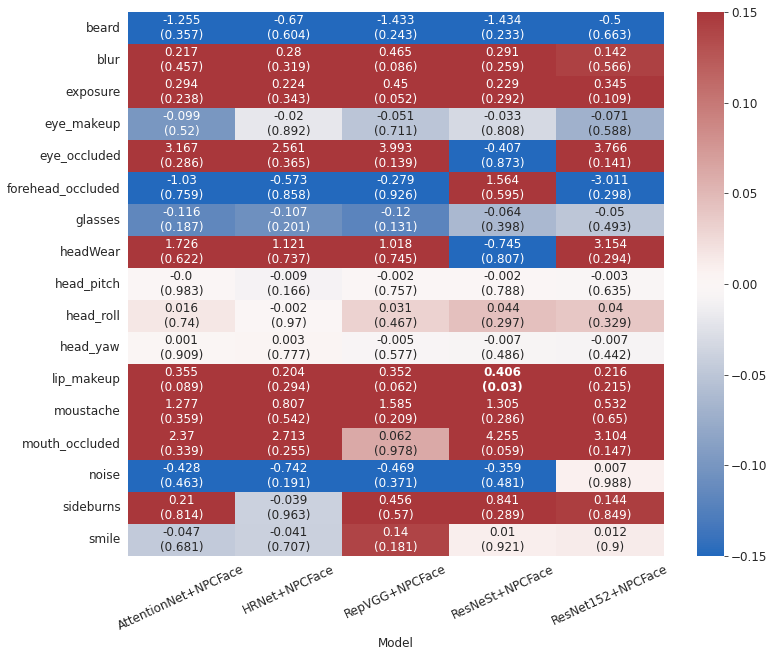}}
\newline
\subfloat[Dependent variable: FRR\label{fig:fine40-tuned}]{
   \includegraphics[width=.65\linewidth, trim=4 4 4 4,clip]{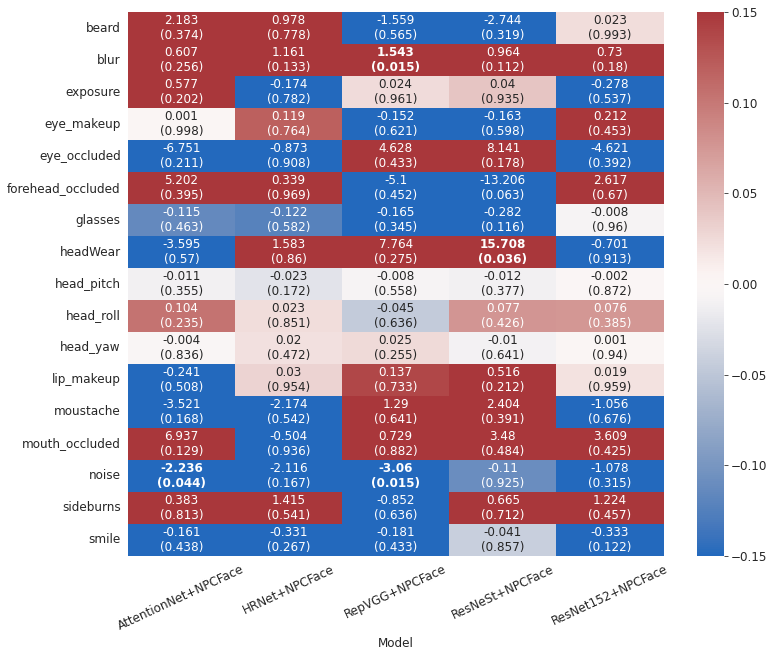}}
\caption{Regression coefficients (p-values) obtained with FAR and FRR as dependent variables on \texttt{VGGFace2}.}
\label{fig:rq2-2}
\end{figure*}

\end{document}